\documentclass[conference]{IEEEtran}
\IEEEoverridecommandlockouts
\usepackage[utf8]{inputenc}
\usepackage{graphicx}
\usepackage{amsmath}
\usepackage[version=4]{mhchem}
\usepackage{siunitx}
\usepackage{amsmath,amsfonts}
\usepackage{subcaption}
\usepackage{mathtools}
\usepackage{booktabs}
\usepackage{multicol}
\usepackage{multirow}
\usepackage{cleveref}
\usepackage{algorithm}
\usepackage{algpseudocode}
\usepackage[backend=biber,style=ieee,sorting=none]{biblatex}
\usepackage{longtable,tabularx}

\begin{document}

\title{Exploring Efficient Quantification of Modeling Uncertainties with Differentiable Physics-Informed Machine Learning Architectures}

\makeatletter
\newcommand{\linebreakand}{%
	\end{@IEEEauthorhalign}
	\hfill\mbox{}\par
	\mbox{}\hfill\begin{@IEEEauthorhalign}
}
\makeatother

\author{
	\IEEEauthorblockN{Manaswin Oddiraju}
	\IEEEauthorblockA{University at Buffalo \\Buffalo, NY, 14260}
	\and
	\IEEEauthorblockN{Bharath Varma Penumatsa}
	\IEEEauthorblockA{University at Buffalo \\Buffalo, NY, 14260}
	\and
	\IEEEauthorblockN{Divyang Amin}
	\IEEEauthorblockA{Bechamo LLC., \\ Buffalo, NY 14260, USA}
	\linebreakand
	\IEEEauthorblockN{Michael Piedmonte}
	\IEEEauthorblockA{Bechamo LLC., \\ Buffalo, NY 14260, USA}
	\and
	\IEEEauthorblockN{Souma Chowdhury }
	\IEEEauthorblockA{University at Buffalo \\ Buffalo, NY, 14260}
}

\maketitle

\begin{abstract}
	Quantifying and propagating modeling uncertainties is crucial for reliability analysis, robust optimization, and other model-based algorithmic processes in engineering design and control. Now, physics-informed machine learning (PIML) methods have emerged in recent years as a new alternative to traditional computational modeling and surrogate modeling methods, offering a balance between computing efficiency, modeling accuracy, and interpretability. However, their ability to predict and propagate modeling uncertainties remains mostly unexplored. In this paper, a promising class of auto-differentiable hybrid PIML architectures that combine partial physics and neural networks or ANNs (for input transformation or adaptive parameter estimation) is integrated with Bayesian Neural networks (replacing the ANNs); this is done with the goal to explore whether BNNs can successfully provision uncertainty propagation capabilities in the PIML architectures as well, further supported by the auto-differentiability of these architectures. A two-stage training process is used to alleviate the challenges traditionally encountered in training probabilistic ML models. The resulting BNN-integrated PIML architecture is evaluated on an analytical benchmark problem and flight experiments data for a fixed-wing RC aircraft, with prediction performance observed to be slightly worse or at par with purely data-driven ML and original PIML models. Moreover, Monte Carlo sampling of probabilistic BNN weights was found to be most effective in propagating uncertainty in the BNN-integrated PIML architectures.
\end{abstract}

%%%%%%%%%%%%%%%%%%%%%%%%%%%%%%%%%%%%%%%%%%%%%%%%%%%%%%%%%%%%%%%%%%%%%%
\section{Introduction}
\label{sec:introduction}
%%%%%%%%%%%%%%%%%%%%%%%%%%%%%%%%%%%%%%%%%%%%%%%%%%%%%%%%%%%%%%%%%%%%%%

Computational frameworks for decision-making in engineering problems often require several iterations of expensive numerical simulations. Surrogates or reduced order models are often used in these circumstances to reduce overall computational cost. However, these models have inherent modeling error and applications in the aerospace domain require quantification of these modeling uncertainties to ensure robust design. Our previous work \cite{oddiraju2023physics,oddiraju2024laser,oddiraju2024physics} has demonstrated the ability of hybrid-physics informed machine learning (PIML) models to act as efficient and interpretable surrogates for a diverse set of engineering applications. There is however limited understanding of model uncertainty quantification choices in such hybrid PIML architectures. In this paper, we therefore investigate a new hybrid-physics-informed machine learning architecture which uses probabilistic neural networks with the goal to achieve fast and efficient quantification of end-to-end modeling uncertainty. The new architectures are tested over an analytical benchmark function, in comparison to pure neural network and PIML baselines.
We then apply this framework to model the aerodynamics of a fixed-wing aircraft and test the accuracy of this model by comparing it with existing Monte Carlo methods. The rest of this section briefly covers existing literature on physics-informed machine learning and uncertainty quantification approaches before stating our research objectives.

\subsection{Physics Informed Machine Learning}

Physics-informed machine learning (PIML) systematically integrates machine learning (ML) algorithms with abstract mathematical models and physical constraints developed within the scientific and engineering domains. Contrary to purely data-centric methods, PIML models can be enhanced by using additional data derived from physical principles such as the conservation of mass and energy. Moreover, abstract characteristics and constraints like stability, convexity, or invariance can be embedded into PIML models. The core concept of PIML is that the fusion of ML and physics leads to the creation of models that are more efficient, consistent with physical laws, and data efficient.

There has been growing interest in integrating data-driven machine learning (ML) approaches with physics-based models, resulting in hybrid frameworks known as Physics-Informed Machine Learning (PIML). These models offer several advantages, including improved generalization, better extrapolation capabilities, enhanced predictive accuracy with limited datasets, and increased interpretability \cite{karniadakis2021physics, rai2020driven}. The structure of PIML frameworks varies depending on the available physics information and the specific objectives of the application.

One widely used approach is the Physics-Informed Neural Network (PINN), which incorporates known partial differential equations (PDEs) directly into the loss function as additional constraints \cite{mao2020physics, raissi2019physics, zhang2022analyses}. PINNs share a similar architecture with conventional neural networks and typically incur comparable computational costs during training and inference, while the physics-based loss term often leads to improved performance \cite{cai2021physics}.

Alternative PIML strategies include the use of physics-inspired activation functions \cite{jagtap2020adaptive, faroughi2023physics} or network architectures specifically designed to learn governing physical laws \cite{li2020fourier, lu2021learning}. For example, Peng et al. applied a convolutional neural network (CNN) augmented with physics-informed loss functions to predict the lift coefficient of an airfoil. Their results demonstrated that incorporating physics into the loss function improved predictive accuracy compared to conventional neural networks.

An emerging category of PIML methods, often termed "Hybrid" PIML \cite{cuomo2022scientific,rai2020driven}, bears some resemblance to graybox modeling \cite{yang2017investigating,manoharan2019grey}. This approach integrates a simplified (typically fast but low-fidelity) physics model component with machine learning model component(s) in varied formats to enhance accuracy or enhance generalization. With the incorporation of physics, their predictions align with physical principles within specific boundaries dictated by their architecture. These architectures are generally categorized into serial \cite{young2017physically,nourani2009combined,singh2019pi} and parallel \cite{javed2014robust,cheng2009fusion,karpatne2017physics} configurations. In \cite{oddiraju2023physics}, the aerodynamic coefficients of a six-rotor eVTOL aircraft are predicted using two PIML modeling methods, factoring in both flight conditions and control inputs. These PIML models are trained and assessed against a purely data-driven ANN and a baseline physics model, demonstrating superior performance by the PIML models. Serial architectures typically arrange data-driven models sequentially with the simplified physics model or use them to dynamically adjust (based on inputs) the parameters or terms of the simplified physics model that might otherwise be fixed or empirically estimated. Parallel architectures tend to employ additive or multiplicative ensembles combining simplified physics models and data-driven ML models.%Parallel models are usually easier to train as the backpropagation to train the neural networks need not go through the physics model. 
In hybrid models, especially those characterized by sequential information flow, a significant challenge lies in ensuring robust and efficient differentiability of the component physics model(s). This is crucial for calculating the gradients of the loss function concerning the parameters of the machine learning model under optimization. Consequently, it necessitates a mechanism for back-propagating through the physics model.

\subsection{Uncertainty Quantification}

Uncertainty quantification (UQ) assesses the reliability of model predictions by considering uncertainties in input parameters, model structure, and data. Aerospace systems must operate safely under extreme conditions. Uncertainty Quantification helps identify and mitigate risks by quantifying uncertainties in material properties, environmental conditions, and system performance. Prototyping and testing in aerospace are expensive. Uncertainty Quantification reduces the need for excessive physical testing by providing confidence in simulation-based designs, cutting costs, and managing development risks. Uncertainty Quantification supports compliance with stringent safety standards by providing credible risk analyses and reliability metrics. So by integrating uncertainty propagation in PIML models, we can reduce the cost and time of physical experiments and also the computational cost of high-fidelity simulations and also fairly remain physically consistent compared to pure data-driven models. \\

Uncertainty propagation approaches can be primarily divided into deterministic \cite{chen2002interval, degrauwe,abdo2016uncertainty} and probabilistic approaches \cite{zhang2021modern,olivier2021bayesian,farid2022data}. Deterministic approaches such as Interval Analysis \cite{chen2002interval,degrauwe} and Fuzzy Theory \cite{abdo2016uncertainty} are quick, computationally cheap, and are good with sparse data but suffer with the complexity of system and accuracy. Though probabilistic approaches like Monte-Carlo Simulations \cite{zhang2021modern}, Bayesian Inference \cite{olivier2021bayesian} and Gaussian Processes \cite{farid2022data} are good with complex systems and are more accurate, they are computationally expensive. In machine learning, uncertainty propagation can be divided into traditional machine learning and deep learning. Gaussian Process Regression (GPR) \cite{farid2022data} and GPR with Physics-Informed Neural Network \cite{pfortner2022physics} are approaches in traditional machine learning while Bayesian Neural Network \cite{olivier2021bayesian} and Bayesian Neural Network with Physics-Informed Neural Network \cite{molnar2022flow,ceccarelli2021bayesian} are approaches in deep learning. Some of the works in the literature on uncertainty propagation in PIML models in recent years are \cite{ mumpower2022physically,mahadevan2022uncertainty,kapusuzoglu2021information,hao2023physics,xu2023physics,fuks2020physics,GAO2023107361}. In \cite{GAO2023107361}, they combined the back propagation neural network and Gaussian process regression to predict multi-axial fatigue life and quantify the uncertainty in aerospace structures. The frozen streamlines (FROST) method, which relies on Monte Carlo Simulations, is used to quantify uncertainty for different PIML models to estimate flow properties for multiphase transport in porous media in \cite{fuks2020physics}. In the article \cite{hao2023physics}, a global sensitivity analysis with PIML framework is performed to identify key characteristic parameters of notch fatigue, while also estimating the lifetime uncertainty induced by notch geometry. Bayesian Neural Networks \cite{goan} are an extension of traditional neural networks that incorporate Bayesian probability theory for uncertainty quantification, model robustness, and decision-making. BNNs are used in Uncertainty estimation \cite{kendall}, Bayesian optimization \cite{shahriari}, and Reinforcement learning \cite{depeweg}. The advantage of BNNs is that they can provide meaningful predictions and quantify uncertainty even with for complex systems. However, uncertainty quantification with BNN is computationally more expensive compared to their non-Bayesian counterparts and also requires certain amount of data to achieve meaningful uncertainty estimates. BNNs have additional hyper-parameters associated with the priors, and tuning them effectively can be challenging. Poorly chosen hyper-parameters may lead to suboptimal performance.

\subsection{Research Objectives}
This brings us to the key objectives and contributions of this paper, which can be summarized as:
\begin{enumerate}
	\item Integrate BNNs into our existing auto-differentiable hybrid PIML architectures that serve in input transformation and adaptive parameter estimation roles, to enable predicting both the mean and uncertainty (e.g., variance or confidence interval) value of the output(s).
	\item Analyze the prediction accuracy and the uncertainty quantification achieved with the BNN-integrated PIML architectures for an analytical benchmark problem, and how they compare with those provided by the simplified physics, the original PIML architecture, and other baselines such as pure data-driven ANN and BNN models.
	\item Apply the new BNN-integrated PIML architecture to real-world flight experiment data and compare its accuracy and uncertainty predictions to relevant existing baselines.
\end{enumerate}

The rest of this paper is laid out as follows: Section \ref{sec:Framework} contains details of our proposed PIML framework. Section \ref{sec:case_study} lists out the case studies in detail. We present our results on these two case studies in Section \ref{sec:Results} before finally ending by stating our conclusions in Section \ref{sec:conclusion}.

\begin{figure*}[]
	\centering
	\includegraphics[width=0.7\linewidth]{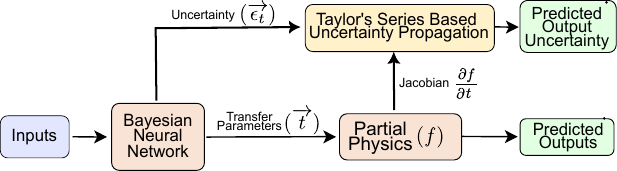}
	\caption{Probabilistic PIML Framework For Efficient Propagation of Modeling Uncertainties}
	\label{fig:uncert_prop}
\end{figure*}
\begin{figure*}[h!]
	\centering
	\includegraphics[width=\linewidth]{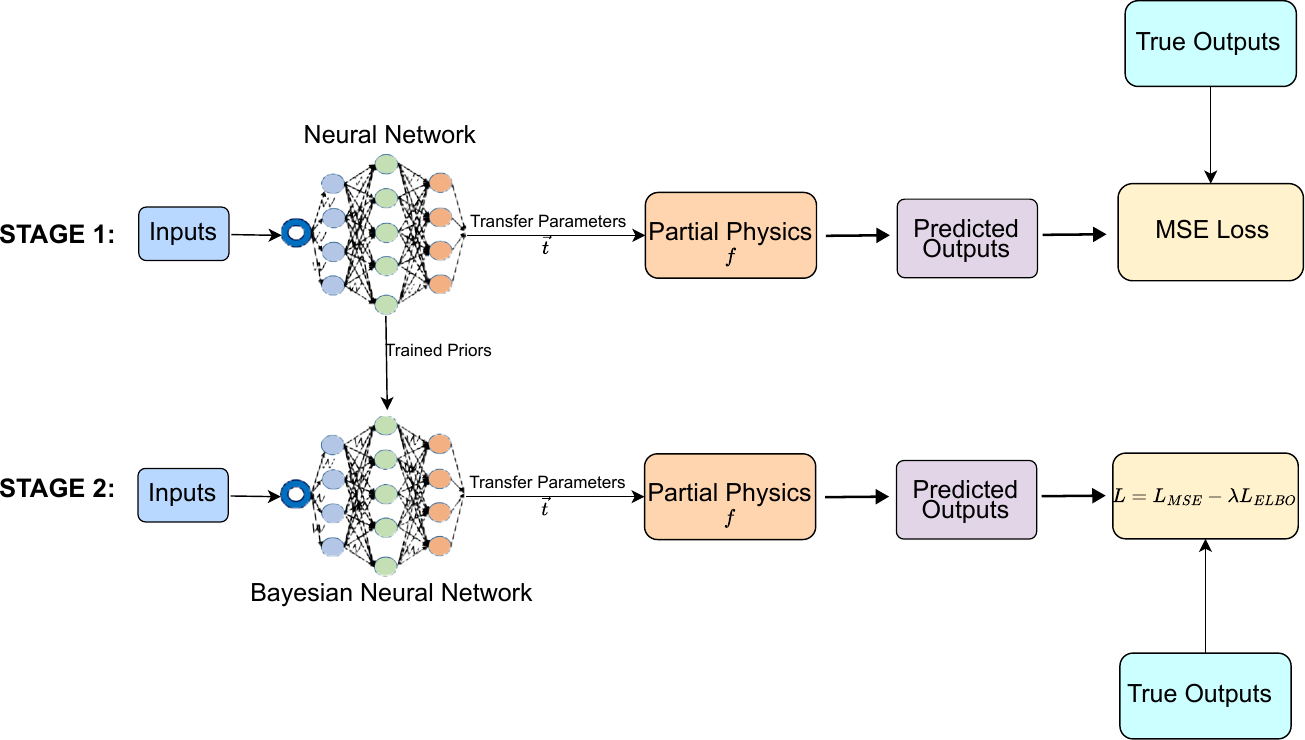}
	\caption{Training of Probabilistic PIML Framework}
	\label{fig:training}
\end{figure*}
%%%%%%%%%%%%%%%%%%%%%%%%%%%%%%%%%%%%%%%%%%%%%%%%%%%%%%%%%%%%%%%%%%%%%%
\section{Physics-Informed Framework for Propagating Modeling Uncertainties}
\label{sec:Framework}
%%%%%%%%%%%%%%%%%%%%%%%%%%%%%%%%%%%%%%%%%%%%%%%%%%%%%%%%%%%%%%%%%%%%%%

The modeling uncertainty (or the uncertainty from data) in our PIML architecture is primarily due to the embedded neural networks. Hence, in order to first quantify and later propagate these uncertainties through the rest of the PIML model, we replace the ANNs from previous deterministic PIML architectures \cite{oddiraju2023physics} with Bayesian Neural Networks.

Figure~\ref{fig:uncert_prop} shows our proposed PIML architecture. The modeling uncertainties here are captured by the Bayesian Neural Network (BNN), which serves to learn intermediate parameter representations or ``transfer parameters''. These parameters are the inputs or parameters of simplified physics models which are learned by the BNN to improve accuracy in the physics models. This section gives a brief overview into our BNN implementation, training, and uncertainty propagation strategies.

\subsection{Bayesian Neural Networks (BNN)}
\label{ssec:BNNs}
%%%%%%%%%%%%%%%%%%%%%%%%%%%%%%%%%%%%%%%%%%%%%%%%%%%%%%%%%%%%%%%%%%%%%%
In the Bayesian framework, probabilities are assigned to events based on prior knowledge or beliefs. We represent the weights and biases as probability distributions, and during inference, a Monte Carlo prediction is run by sampling over the weight distribution to get the distribution of the outputs. A Bayesian Neural Network can be represented mathematically as:
\begin{equation}
	p(\hat{y}(x) \mid D)=\int_w p(\hat{y}(x) \mid w) p(w \mid D) d w=\mathbb{E}_{p(w \mid D)}[p(\hat{y}(x) \mid w)]
\end{equation}
Here, $p(y|x,\theta)$ is the likelihood function of $y$ given the input data set ($x$), and weights and biases ($\theta$), $p(\theta|D)$ is the posterior distribution of weights and biases given the observed data $D$.

BNNs, however useful in theory, are notoriously difficult to train and often have lackluster prediction accuracy. Therefore, to preserve prediction accuracy while adding the ability to quantify uncertainties, we apply the Bayesian modeling paradigm only to the last layer of the neural network. This limits the number of probabilistic weights to the output layer, which helps make training easier.

\subsection{Training the Probabilistic PIML Model}
\label{ssec:Training}
%%%%%%%%%%%%%%%%%%%%%%%%%%%%%%%%%%%%%%%%%%%%%%%%%%%%%%%%%%%%%%%%%%%%%%%%%%

Traditional training approaches for BNNs either depend on sampling-based methods \cite{green}, or more popularly on Stochastic Variational Inference (SVI) \cite{hoffman2013stochastic}.
In this paper, to train our BNN-PIML architecture, we use a two-stage training procedure, as shown in Fig.~\ref{fig:training}. First, we train the PIML architecture using traditional ANNs with an MSE loss. This allows for fast training and rapid iterations necessary for hyper-parameter optimization. Once the deterministic PIML architecture is trained, we use the learned weights as priors for the BNN and retrain it using a weighted combination of MSE and ELBO losses. This two-stage approach sidesteps the challenges typically faced when training BNNs and results in a more accurate PIML architecture capable of predicting uncertainties.

During training, only the weights of the final layer are treated as probabilistic variables under the Bayesian assumption, while the remaining layers are deterministic. Therefore, the total loss combines the standard MSE loss with the Bayesian ELBO loss:

\begin{equation}
	\begin{aligned}
		\mathcal{L} = \mathcal{L}_{\mathrm{MSE}} - \lambda \mathcal{L}_{\mathrm{ELBO}}
	\end{aligned}
\end{equation}
where $\lambda$ is a hyperparameter that controls the contribution of the Bayesian loss. The prior distribution for the Bayesian weights is initialized using the mean and standard deviation of the corresponding weights from the previously trained deterministic models. This informed initialization greatly improves convergence compared to starting with random priors.

Fig. \ref{fig:uncert_prop} shows the framework for uncertainty propagation in the PIML-BNN model. In the framework, we used vector notation to represent the parameters as vectors. The inputs are passed to the Bayesian neural network which gives transfer parameters and uncertainty in them. Transfer parameters are given as input to partial (low-fidelity) physics model which gives predicted outputs and Jacobian. Jacobian and Uncertainty in transfer parameters are then used to approximate output uncertainty.

\subsection{Uncertainty Propagation through PIML Models}
%%%%%%%%%%%%%%%%%%%%%%%%%%%%%%%%%%%%%%%%%%%%%%%%%%%%%%%%%%%%%%%%%%%%%%

As Hybrid PIML models are differentiable, additive uncertainties in the transfer parameters, can be propagated efficiently by using a first order Taylor's series expansion as shown in Eq.~\ref{eq:Uncertainty_1}.

\begin{equation}
	\label{eq:Uncertainty_1}
	\epsilon_{y_j} = \Sigma_{i=1}^p \dfrac{\partial f_j}{\partial t_i} \epsilon_t^i \qquad    \text{ ,for j = 1,2,3...m }
\end{equation}

Here, $\epsilon_{y_j}$ represents the uncertainty of the $j^{\rm{th}}$ output, $t_i$ and $\epsilon_{t}^i$ are the value and uncertainty of the $i^{\rm{th}}$ transfer parameter respectively.

This approach replaces Monte Carlo simulations which are usually required for uncertainty propagation thereby improving computational efficiency of the overall uncertainty modeling framework.

% This uncertainty can be modeled as additive noise, and the output error of the differentiable physics model is computed using Taylor's series approximation as shown in Eq.[\ref{eq:Uncertainty_2}]. Here, the derivative terms from the second order are ignored.

% Where $\epsilon_y$ is the output error and $m$ is the number of outputs of the partial physics model.

\section{Case Studies}
\label{sec:case_study}
%%%%%%%%%%%%%%%%%%%%%%%%%%%%%%%%%%%%%%%%%%%%%%%%%%%%%%%%%%%%%%%%%%%%%%

To test the performance of the probabilistic PIML model, we use a combination of analytical and real-world test cases. The rest of this section contains the details of both these test cases.

\subsection{Analytical Case Study: Gramacy \& Lee Function}
\label{ssec:gramacy_lee}
%%%%%%%%%%%%%%%%%%%%%%%%%%%%%%%%%%%%%%%%%%%%%%%%%%%%%%%%%%%%%%%%%%%%%%

We used a highly multi-modal variation of the Gramacy $\&$ Lee function. We consider one full physics model, $\boldsymbol{f}_{\mathrm{FP}}(\boldsymbol{x})$, and one partial physics model, $f_{\mathrm{PP}}(\boldsymbol{x})$. We used this variation because it has a simple input correlation between the partial physics and the full physics model, and it being a 1-D model will help us visualize it better compared to multi-dimensional models. We can map the partial physics model to the full physics model using an input transfer,r after which the partial physics model completely matches the full physics model. The ideal input transfer is respectively:
$$
	\chi_T =0.5+2 \sin (\pi(x-0.5) / 4)
$$

The partial physics model and the full physics model expressions and mapping of the partial physics model to the full physics model using ideal input transfer parameter is given below in Eq. \ref{eq:gl}

\begin{equation}\label{eq:gl}
	\begin{aligned}
		f_{\mathrm{PP}}(x) & = \frac{\sin (10 \pi x)}{2 x}+(x-1)^4 ; x \in[0.5,2.5]                 \\
		f_{\mathrm{FP}}(x) & = \frac{\sin (10 \pi(0.5+2 \sin (\pi(x-0.5) / 4)))}{2(\pi(x-0.5) / 4)} \\
		                   & + ((0.5+2 \sin (\pi(x-0.5) / 4))-1)^4                                  \\
		f_{\mathrm{FP}}(x) & = f_{\mathrm{PP}}(0.5+2 \sin (\pi(x-0.5) / 4))
	\end{aligned}
\end{equation}
\begin{figure}[h!]
	\centering
	\includegraphics[width=\linewidth]{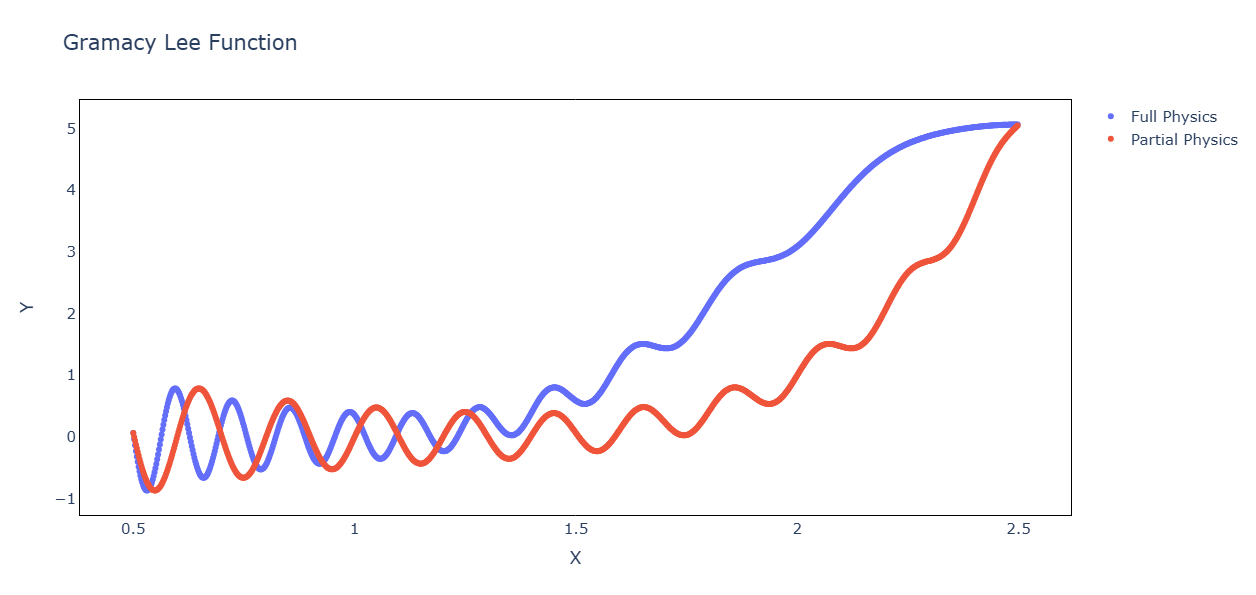}
	\caption{Partial (Low-fidelity) and Full Physics (High-fidelity) Models of Gramacy $\&$ Lee Function}
	\label{fig:gl}
\end{figure}
Fig. \ref{fig:gl} shows the $f$ vs. $x$ plots for the partial physics and the full physics model of the Gramacy $\&$ Lee problem. We can see the output space correlation between the partial physics model and the full physics model.

\begin{figure}
	\centering
	\includegraphics[width=.8\linewidth]{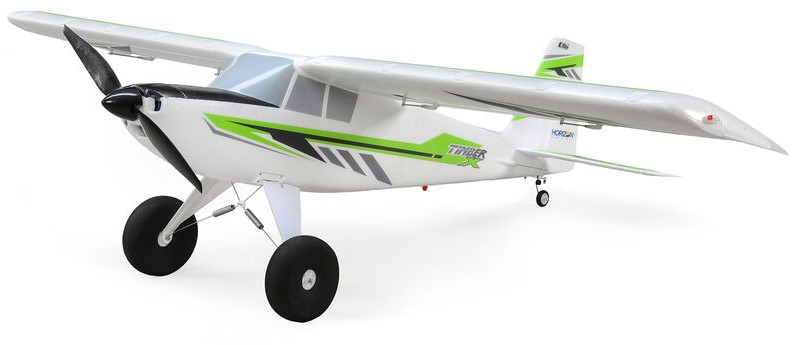}
	\caption{Image of a Timber-X model aircraft, used as our second case study.}
	\label{fig:Timber}
\end{figure}

\subsection{Aerodynamics of a Fixed Wing Aircraft}
\label{ssec:case_study_aircraft}
%%%%%%%%%%%%%%%%%%%%%%%%%%%%%%%%%%%%%%%%%%%%%%%%%%%%%%
In this case study, we aim to predict the force components of the aircraft\footnote{https://www.amainhobbies.com/eflite-timber-x-1.2m-pnp-electric-airplane-1200mm-efl3875/p891554}. Our inputs are free stream velocity, angle of attack ($\alpha$), sideslip angle ($\beta$), aileron angle, rudder angle, and the throttle. The transfer parameters are chosen the same as the inputs and help in scaling the inputs from normalized data.

\subsubsection{Partial (Low-fidelity) Physics Model}
% Our reference physics model shown in Fig.~\ref{fig:aerobeacon_lf}, 
The lower fidelity flight dynamics model uses Bechamo LLC's developed VLM plus propeller effects low-fidelity tool, BLOFI.  Implementation and validation of the tool are cited in \cite{oddiraju2023physics}. The source of the high-fidelity data is obtained from flight tests. The experimental nature of the data itself raises additional observability challenges as the aerodynamic coefficients cannot be recorded during experimentation, and therefore, our outputs are the aerodynamic forces. To account for this change in the modeling quantities of interest (QOI), we modify our low-fidelity physics model as well as our physics-informed architectures accordingly, the details of which are provided in this section. We get the aerodynamic coefficients in the wind frame ($C_L, C_D, C_Y$) from the VLM implementation. These coefficients are then converted to body frame coefficients using the Eq. \ref{eq:wind_inertial}. These equations are simplified equations of conversion under the assumption that the side slip angle is very small. We verified this with the experimental data and found that $\beta$ is indeed small and is in the range $[-2^o,10^o]$.

\begin{equation}
	\begin{aligned}
		 & \mathcal{C}_{\mathcal{X}}=-\mathrm{C}_{\mathrm{D}} \cos \alpha+\mathrm{C}_{\mathrm{L}} \sin \alpha \\
		 & \mathcal{C}_{\mathcal{Y}}=\mathrm{C}_{\mathrm{Y}}                                                  \\
		 & \mathcal{C}_{\mathcal{Z}}=-\mathrm{C}_{\mathrm{D}} \sin \alpha-\mathrm{C}_{\mathrm{L}} \cos \alpha
	\end{aligned}
	\label{eq:wind_inertial}
\end{equation}
Where $C_L, C_D \ \text{and} \ C_Y$ are aerodynamic lift, drag, and side force coefficients. $C_X, C_Y \ \text{and} \ C_Z$ are force coefficients in body frame and $\alpha$ is the angle of attack. After converting to body frame coefficients, we used Eq. \ref{eq:aero_comp} to calculate body frame aerodynamic force components.

\begin{equation}
	\begin{aligned}
		 & \mathrm{F}_{\mathrm{X}_\mathrm{aero}}= \mathrm{q}_{\infty}\mathrm{S}_{\mathrm{ref}}\mathcal{C}_{\mathcal{X}} \\
		 & \mathrm{F}_{\mathrm{Y}_\mathrm{aero}}= \mathrm{q}_{\infty}\mathrm{S}_{\mathrm{ref}}\mathcal{C}_{\mathcal{Y}} \\
		 & \mathrm{F}_{\mathrm{Z}_\mathrm{aero}}= \mathrm{q}_{\infty}\mathrm{S}_{\mathrm{ref}}\mathcal{C}_{\mathcal{Z}}
	\end{aligned}
	\label{eq:aero_comp}
\end{equation}

Where $\mathrm{q}_{\infty} = \frac{1}{2}\rho{\mathrm{V}_\infty}^2$ is the dynamic pressure, $\rho$ is the density of air, $V_{\inf}$ is the free stream velocity. $S_{\text{ref}}$ is the reference surface area (wing surface area in the case of planes). $\mathrm{F}_{\mathrm{X}_\mathrm{aero}}, \mathrm{F}_{\mathrm{Y}_\mathrm{aero}} \ \text{and} \ \mathrm{F}_{\mathrm{Z}_\mathrm{aero}}$ are aerodynamic force components in body frame.

We get the propeller force components ($F_{prop}$) from the prop solver in VLM implementation. Adding up the aerodynamic force components ($F_{aero}$), propeller force components ($F_{prop}$) and force due to gravity ($F_{gravity}$), we get the net forces as given in Eq. \ref{eq:forces}.

\begin{equation}
	\mathrm{F}_{\mathrm{net}} = \mathrm{F}_{\mathrm{aero}} + \mathrm{F}_{\mathrm{prop}} + \mathrm{F}_{\mathrm{gravity}}
	\label{eq:forces}
\end{equation}

\subsubsection{High-Fidelity Dataset}

We use flight test data as the source of our high-fidelity data. The temperature and humidity for this flight data are close to standard day, and the wind was very low. We only considered the data when the control surfaces (rudder, aileron, and elevator) were active. The flight data has 6 inputs mentioned above and 3 force components as outputs. The flight data is collected from 3 different flight configurations. They are steady climb, varied elevator trim, and steady altitude. The flight data samples are taken such that there is uniform coverage of the domain of the input variables.

% \begin{figure}
%     \centering
%     \includegraphics[width=0.65\linewidth]{Figures/monty_ann.pdf}
%     \caption{ANN Framework for Fixed Wing Aircraft}
%     \label{fig:ann_fw_monty}
% \end{figure}

% \begin{figure}
%     \centering
%     \includegraphics[width=1\linewidth]{Figures/monty_piml.pdf}
%     \caption{PIML-ANN Framework for Fixed Wing Aircraft}
%     \label{fig:pimlann_fw_monty}
% \end{figure}

% \begin{figure}
%     \centering
%     \includegraphics[width=0.7\linewidth]{Figures/monty_bnn.pdf}
%     \caption{BNN Framework for Fixed Wing Aircraft}
%     \label{fig:bnn_fw_monty}
% \end{figure}

% \begin{figure}
%     \centering
%     \includegraphics[width=1\linewidth]{Figures/monty_pimlbnn.pdf}
%     \caption{PIML-BNN Framework for Fixed Wing Aircraft}
%     \label{fig:pimlbnn_fw_monty}
% \end{figure}
%  Fig. \ref{fig:ann_fw_monty} and \ref{fig:pimlann_fw_monty} show the frameworks of pure data-driven ANN and PIML-ANN models for the fixed-wing aircraft study case. Fig. \ref{fig:bnn_fw_monty} and \ref{fig:pimlbnn_fw_monty} show the frameworks of pure data-driven BNN and PIML-BNN models for the fixed-wing aircraft study case.

\section{Results and Discussion}
\label{sec:Results}
%%%%%%%%%%%%%%%%%%%%%%%%%%%%%%%%%%%%%%%%%%%%%%%%%%%%%%%%%%%%%%%%%%%%%%

This section outlines the performance of the trained PIML-BNN model on both the above detailed test cases. To get a fair estimate of performance, we also use a traditional ANN, PIML-ANN, and a traditional BNN as baselines. Tab.~\ref{tab:config} \& Tab.~\ref{tab:config1} list the model architectures across case studies respectively. Training convergence histories for both case studies are provided in Appendix B.

\subsection{Modeling Results on the Analytical Case Study}

\begin {table}[h!]
\centering
\resizebox{\columnwidth}{!}{%
	\begin{tabular}{|c|c|c|c|c|}
	% \toprule[1.5pt]
	\hline
	{\bf } & {\bf PIML-ANN} & {\bf ANN} & {\bf PIML-BNN} & {\bf BNN}\\
	% \midrule
	\hline
	Hidden Layers        & 5     & 5      & 5     & 5 \\
	\hline
	Nodes per Layer        & 200     & 200      & 200     & 200 \\
	\hline
	Learning Rate        & $10^{-4}$     & $10^{-4}$      & $10^{-3}$     & $10^{-3}$ \\
	\hline
	Training Samples   & 900     & 900      & 900     & 900 \\
	\hline
	Testing Samples   & 100     & 100      & 100     & 100 \\
	\hline
	Training Time (min)     &  0.72     & 0.6      &  5.5     &  3.2 \\
	\hline
	Epochs        & 1000     & 1000      & 1000     & 1000 \\
	\hline
	Activation Function  &  &  &  &  \\
	(hidden layers)&  Leaky ReLU     & Leaky ReLU      &  Leaky ReLU     &  Leaky ReLU \\
	\hline
	No. of Inputs & 1 & 1 & 1 & 1 \\
	\hline
	No. of Transfer & & & & \\
	Parameters & 1 & - & 1 & - \\
	\hline
	No. of Outputs & 1 & 1 & 1 & 1 \\
	\hline
	% \bottomrule[1.25pt]
	\end {tabular}%
}
\caption{Configuration and Training Times of Models for Gramacy \& Lee problem}
\label{tab:config}

\end {table}

\begin{figure*}
	\centering
	\begin{subfigure}{0.45\linewidth}
		\centering
		\includegraphics[width=\linewidth]{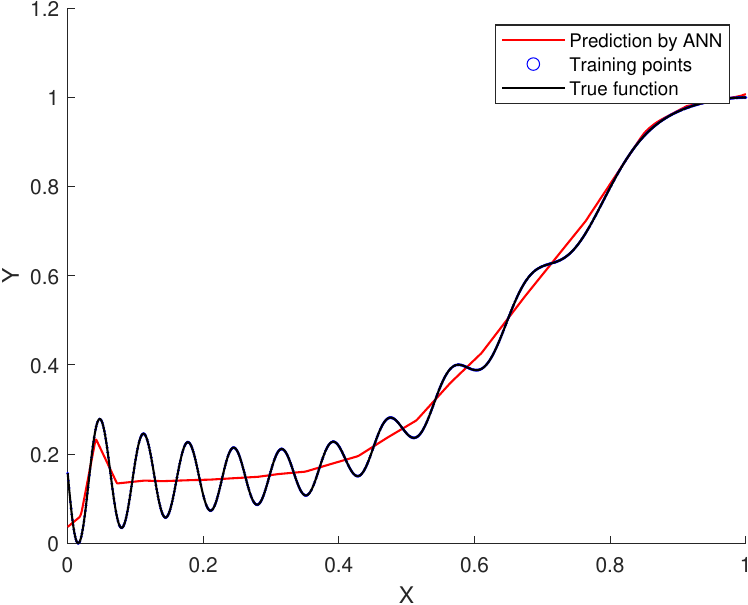}
		\caption{}
		\label{fig:ann_gl}
	\end{subfigure}
	\begin{subfigure}{0.45\linewidth}
		\centering
		\includegraphics[width=\linewidth]{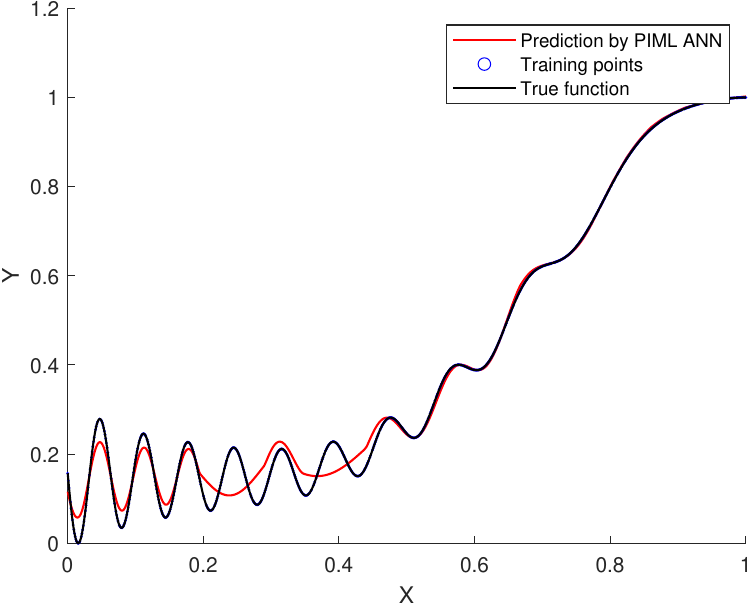}
		\caption{}
		\label{fig:piml_gl}
	\end{subfigure}
	\caption{Output Predictions of: a) ANN, b) PIML-ANN on the Analytical Case Study}
	\label{fig:pred_nn}
\end{figure*}

Fig.\ref{fig:rmse_gl} shows that the PIML-ANN, ANN, and BNN models perform similarly in terms of prediction accuracy, while the PIML-BNN model performs slightly worse. Fig. \ref{fig:pred_nn} shows clearly that PIML-ANN is predicting better compared to pure data-driven ANN. From Fig.\ref{fig:uncert_gl}, it is evident that both the mean and uncertainty predictions deviate significantly from the true function and training points at the start. This may be attributed to how uncertainty is propagated through the BNN transfer layer. To address this, we employed Monte Carlo simulations to estimate uncertainty, leveraging the fact that the weights in the BNN transfer network are modeled as distributions.

We conducted 20 Monte Carlo simulations for models with varying numbers of hidden units (10, 20, 50, and 200) in the transfer network. Models with 10 hidden units converged faster and yielded better predictions compared to those with more hidden units. This is likely due to the relatively simple nature of the problem, where the full physics model is a non-linear function that does not require a highly complex network. While Monte Carlo simulations improved results, they become impractical for more complex problems or large datasets due to increased computational costs.

Fig.~\ref{fig:uncert_gl_mc} shows the mean prediction and uncertainty plots of the BNN and PIML-BNN models using 10 hidden units and 20 Monte Carlo simulations, with only the last layer modeled as Bayesian.

\begin{figure}[h!]
	\centering
	\includegraphics[width=0.9\linewidth]{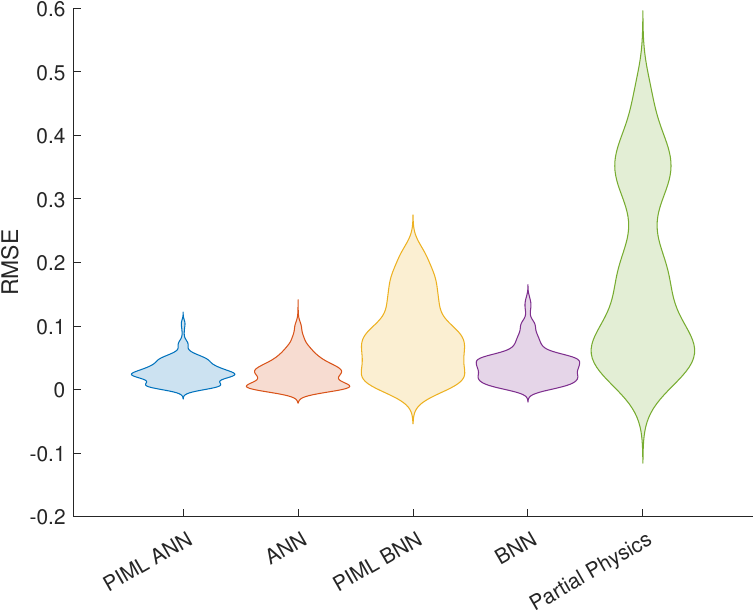}
	\caption{Comparison of Prediction Errors on the Analytical Modeling Case Study}
	\label{fig:rmse_gl}
\end{figure}

\begin{figure*}
	\centering
	\begin{subfigure}{0.45\linewidth}
		\centering
		\includegraphics[width=\linewidth]{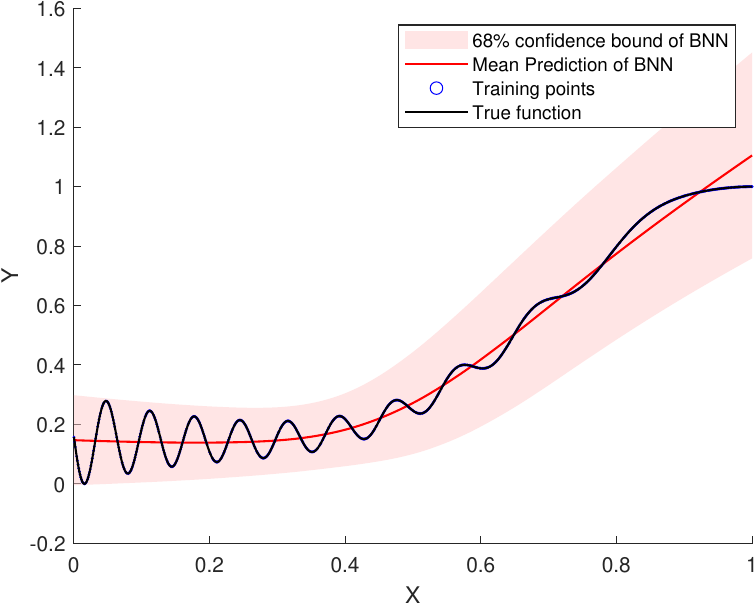}
		\caption{}
		\label{fig:bnn_gl}
	\end{subfigure}
	\begin{subfigure}{0.45\linewidth}
		\centering
		\includegraphics[width=\linewidth]{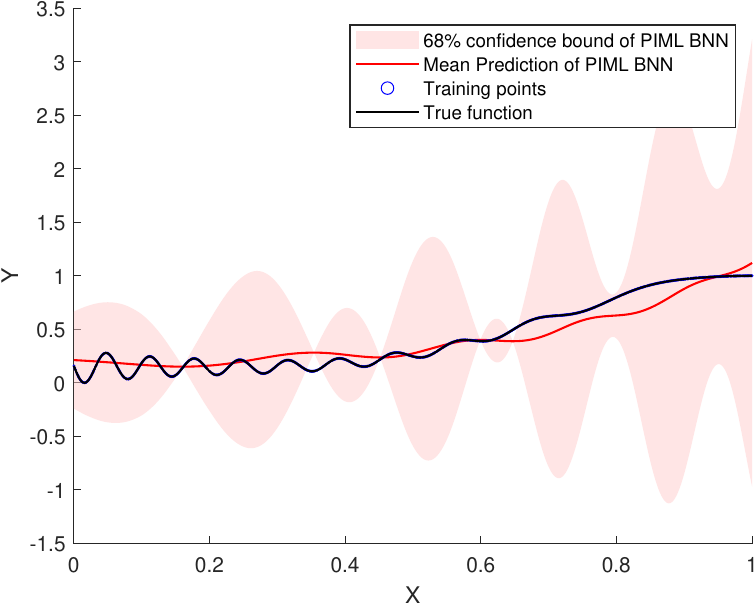}
		\caption{}
		\label{fig:pimlbnn_gl}
	\end{subfigure}
	\caption{Uncertainty Predictions of : a) BNN, b) PIML-BNN, on the Analytical Case Study using internal Monte-Carlo for transfer network and Taylor Series Expansion for the Partial (Low-Fidelity) Physics Model}
	\label{fig:uncert_gl}
\end{figure*}

\begin{figure*}
	\centering
	\begin{subfigure}{0.45\linewidth}
		\centering
		\includegraphics[width=\linewidth]{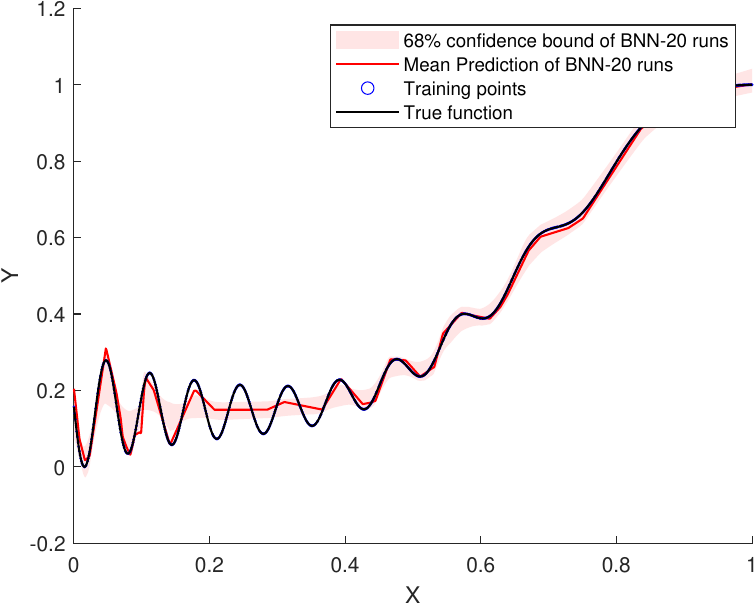}
		\caption{}
		% \label{fig:bnn_gl}   
	\end{subfigure}
	\begin{subfigure}{0.45\linewidth}
		\centering
		\includegraphics[width=\linewidth]{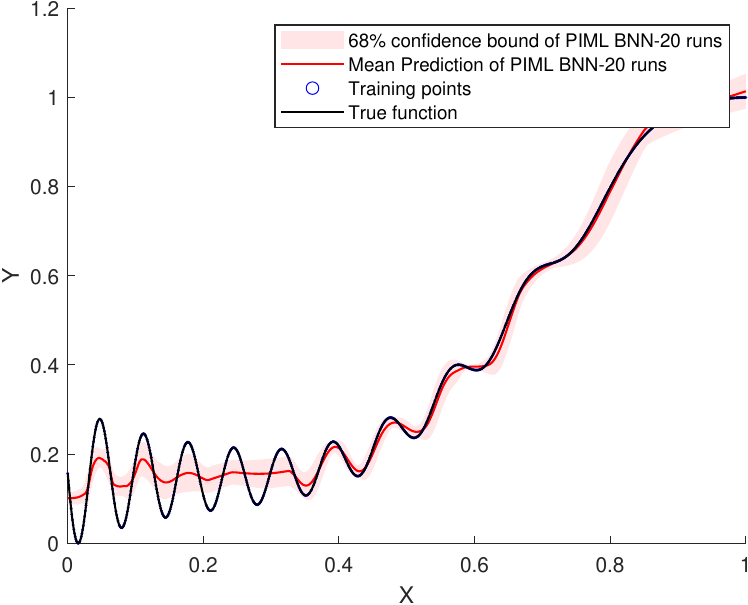}
		\caption{}
		% \label{fig:pimlbnn_gl}   
	\end{subfigure}
	\caption{Uncertainty Predictions of : a) BNN, b) PIML-BNN, on the Analytical Case Study with end-to-end Monte-Carlo simulations}
	\label{fig:uncert_gl_mc}
\end{figure*}

%%%%%%%%%%%%%%%%%%%%%%%%%%%%%%%%%%%%%%%%%%%%%%%%%%%%%%%%%%%%%%%%%%%%%%
\subsection{Aerodynamics of a Fixed Wing RC Aircraft}
Table \ref{tab:config1} shows the configuration and training times of pure data-driven ANN, PIML-ANN, pure data-driven BNN, and PIML-BNN models.
\begin{table}[h!]
\centering
\resizebox{\columnwidth}{!}{%
	\begin{tabular}{|c|c|c|c|c|}
	% \toprule[1.5pt]
	\hline
	{\bf } & {\bf PIML-ANN} & {\bf ANN} & {\bf PIML-BNN} & {\bf BNN}\\
	% \midrule
	\hline
	Hidden Layers        & 5     & 5      & 5     & 5 \\
	\hline
	Nodes per Layer        & 200     & 200      & 200     & 200 \\
	\hline
	Learning Rate        & $10^{-4}$     & $10^{-4}$      & $10^{-3}$     & $10^{-3}$ \\
	\hline
	Training Samples   & 900     & 900      & 900     & 900 \\
	\hline
	Testing Samples    & 100     & 100      & 100     & 100 \\
	\hline
	Training Time (min)     &  157     & 0.16     &  170     &  2 \\
	\hline
	Epochs        & 200     & 200      & 200     & 200 \\
	\hline
	Activation Function  &  &  &  &  \\
	(hidden layers)&  Leaky ReLU     & Leaky ReLU      &  Leaky ReLU     &  Leaky ReLU \\
	\hline
	No. of Inputs & 6 & 6 & 6 & 6 \\
	\hline
	No. of Transfer & & & & \\
	Parameters & 6 & - & 6 & - \\
	\hline
	No. of Outputs & 3 & 3 & 3 & 3 \\
	\hline
	% \bottomrule[1.25pt]
	\end {tabular}%
}
\caption {Configuration and Training Times of Models for Fixed Wing Aircraft Aerodynamics Model}
\label{tab:config1}

\end {table}

To evaluate the performance of the PIML-BNN models, we compared them against the suite of above mentioned baselines. Both the data-driven and PIML-based models were trained on a dataset of 900 high-fidelity samples and validated on a separate set of 100 samples. The priors, $\mu_{\mathrm{prior}}$ and $\sigma_{\mathrm{prior}}$, for the weights in both the stand-alone BNN and PIML-BNN were initialized using the mean and standard deviation of weights from previously trained stand-alone ANN and PIML-ANN models.

Fig.~\ref{fig:rmse_monty} presents violin plots of the RMSE (Root Mean Square Error) for all three force components and the total RMSE for the ANN, PIML-ANN, BNN, and PIML-BNN models. The results show that all models achieve similar prediction accuracy, with the PIML-BNN model exhibiting slightly higher errors overall.

Figs.~\ref{fig:uncert_monty_piml} and \ref{fig:uncert_monty_bnn} further illustrate that the PIML-BNN model underperforms relative to the pure data-driven BNN, except for the $F_y$ component. This discrepancy may stem from limitations in the partial physics (VLM) model, particularly the propeller solver's inability to accurately predict thrust, which directly impacts $F_x$ and $F_z$. Another contributing factor could be the method of uncertainty propagation through the BNN network. Applying Monte Carlo simulations to the entire model could potentially improve predictions but would significantly increase computational cost due to the inclusion of the physics model.

\begin{figure*}
	\centering
	\begin{subfigure}{0.24\linewidth}
		\centering
		\includegraphics[width=\linewidth]{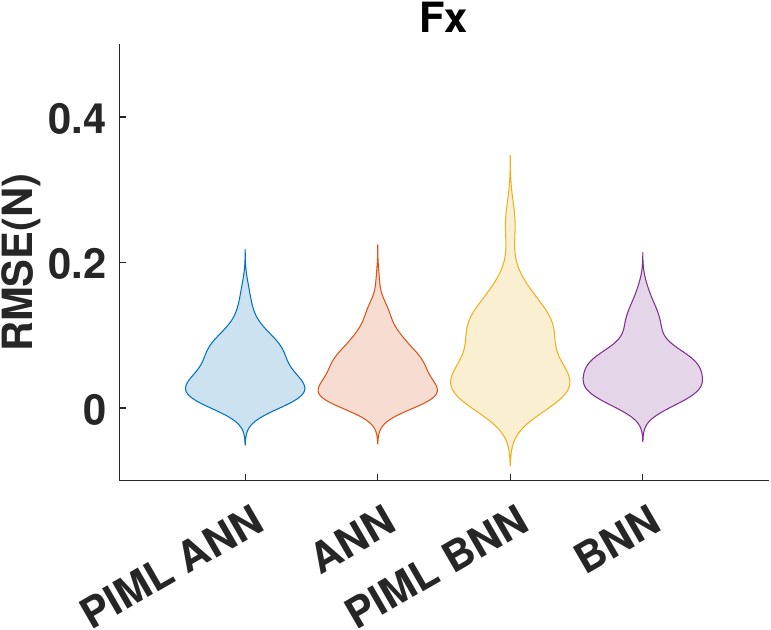}
		\caption{}
		% \label{fig:uncert_fx_piml}
	\end{subfigure}
	\begin{subfigure}{0.24\linewidth}
		\centering
		\includegraphics[width=\linewidth]{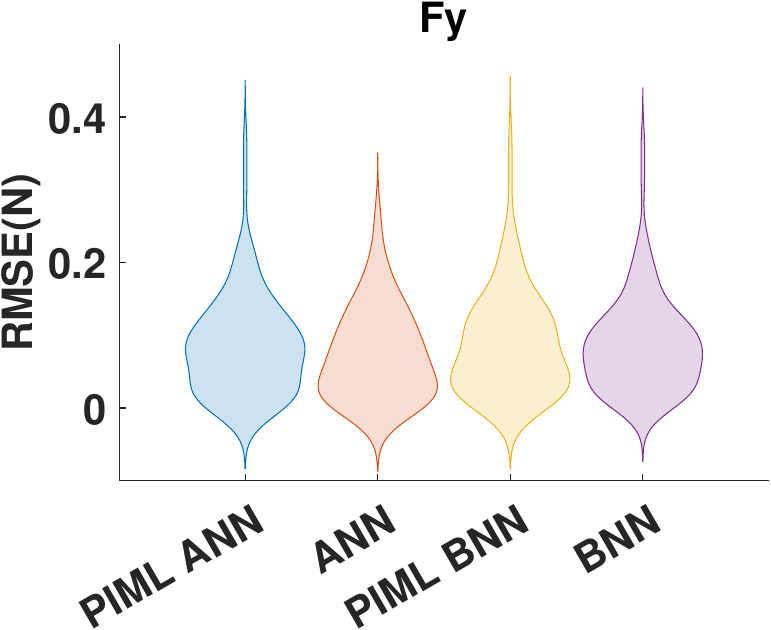}
		\caption{}
		% \label{fig:uncert_fy_piml}
	\end{subfigure}
	\begin{subfigure}{0.24\linewidth}
		\centering
		\includegraphics[width=\linewidth]{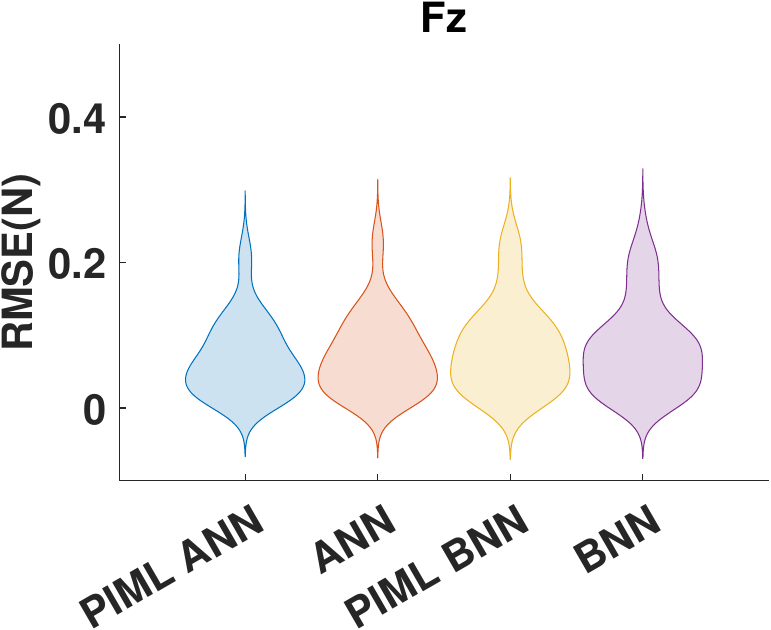}
		\caption{}
		% \label{fig:uncert_fz_piml}
	\end{subfigure}
	\begin{subfigure}{0.24\linewidth}
		\centering
		\includegraphics[width=\linewidth]{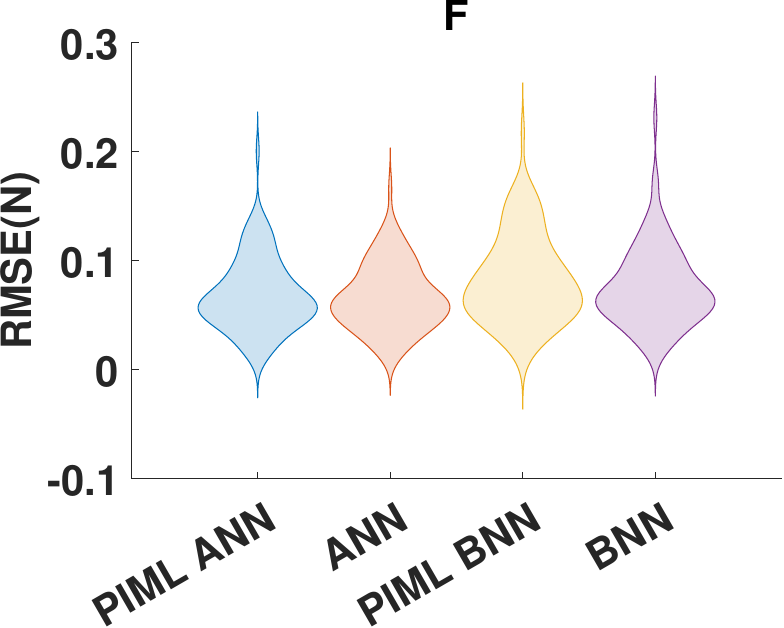}
		\caption{}
		% \label{fig:rmse_monty}
	\end{subfigure}
	\caption{Prediction Error of : a) $F_x$, b) $F_y$, c) $F_z$, d) $F$ (Net Force) on the Fixed Wing Aircraft Case Study}
	\label{fig:rmse_monty}
\end{figure*}

\begin{figure}[]
	\centering
	\begin{subfigure}{0.7\linewidth}
		\centering
		\includegraphics[width=\linewidth]{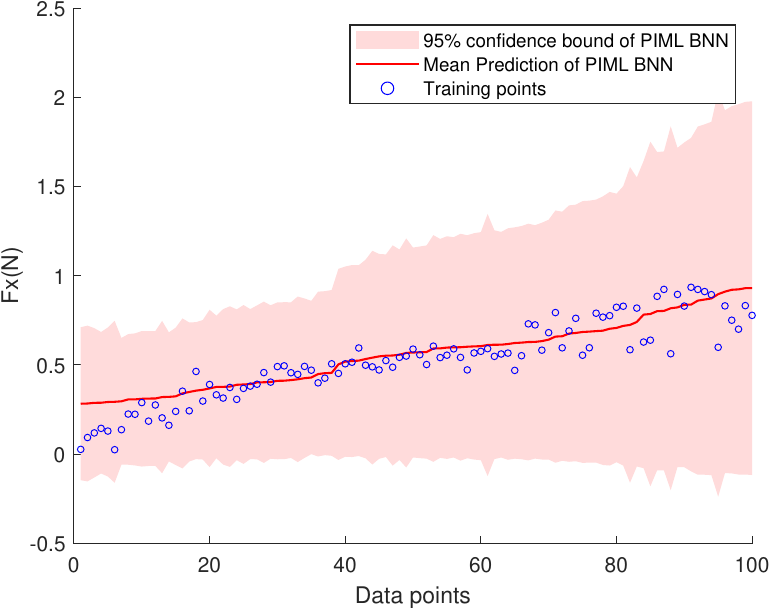}
		\caption{}
		% \label{fig:uncert_fx_piml}
	\end{subfigure}
	\begin{subfigure}{0.7\linewidth}
		\centering
		\includegraphics[width=\linewidth]{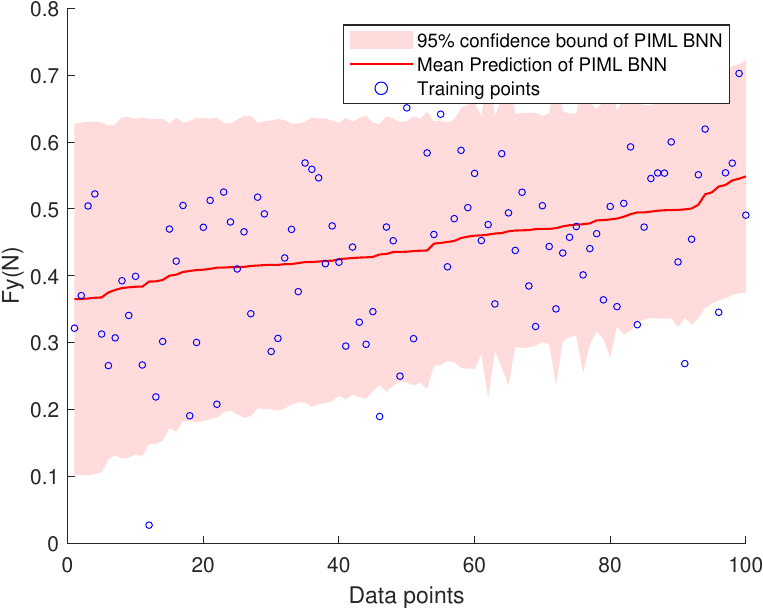}
		\caption{}
		% \label{fig:uncert_fy_piml}
	\end{subfigure}
	\begin{subfigure}{0.7\linewidth}
		\centering
		\includegraphics[width=\linewidth]{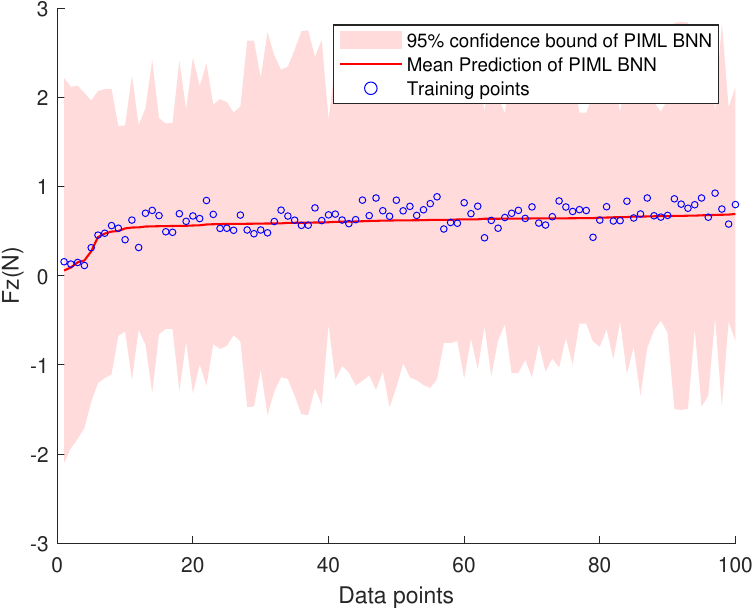}
		\caption{}
		% \label{fig:uncert_fz_piml}
	\end{subfigure}
	\caption{Uncertainty Predictions of: a) $F_x$, b) $F_y$, c) $F_z$, on the Fixed Wing Aircraft Case Study}
	\label{fig:uncert_monty_piml}
\end{figure}

\begin{figure}[h!]
	\centering
	\begin{subfigure}{0.7\linewidth}
		\centering
		\includegraphics[width=\linewidth]{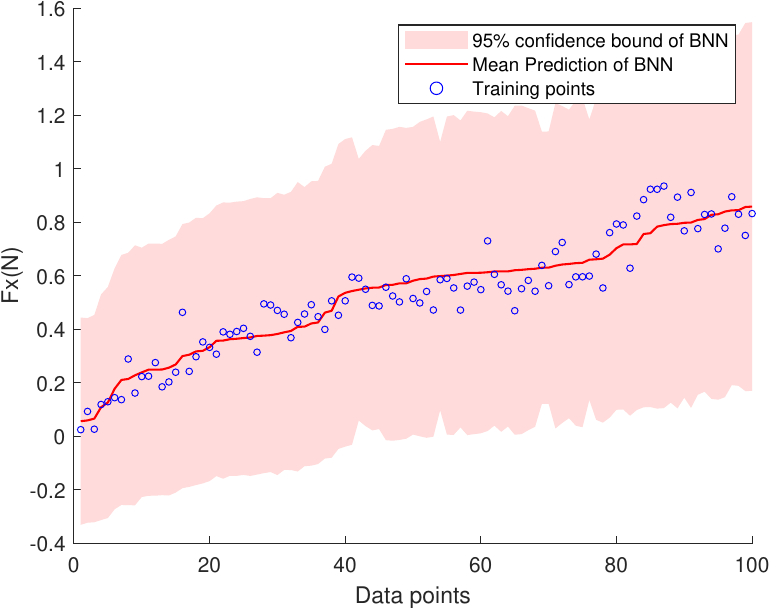}
		\caption{}
		% \label{fig:uncert_fx_bnn}
	\end{subfigure}
	\begin{subfigure}{0.7\linewidth}
		\centering
		\includegraphics[width=\linewidth]{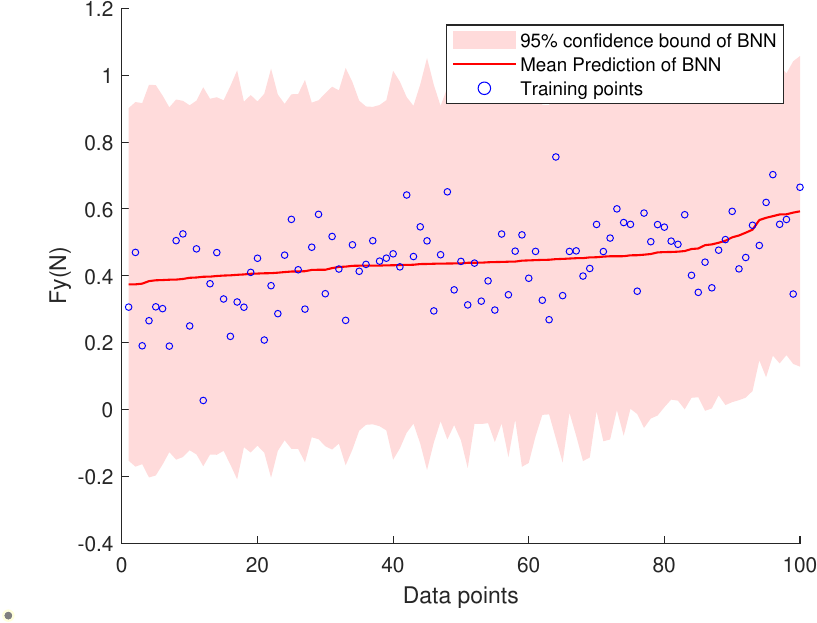}
		\caption{}
		% \label{fig:uncert_fy_bnn}
	\end{subfigure}
	\begin{subfigure}{0.7\linewidth}
		\centering
		\includegraphics[width=\linewidth]{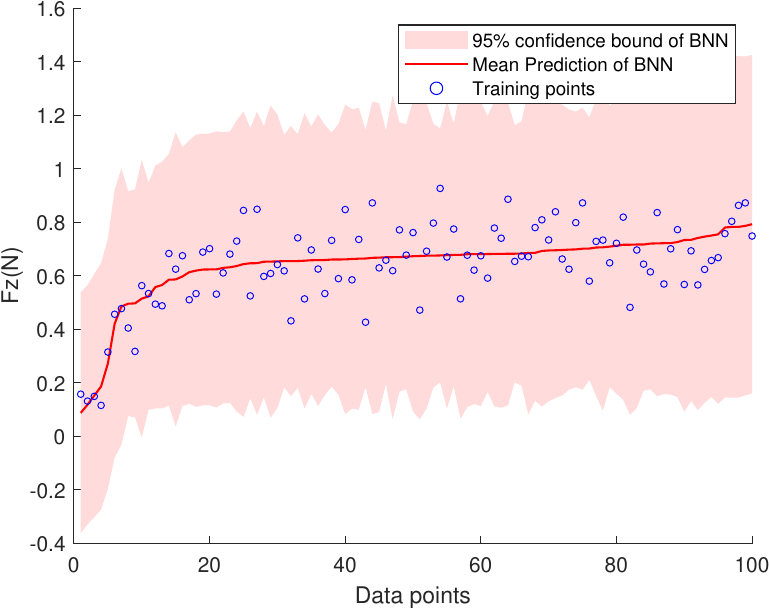}
		\caption{}
		% \label{fig:uncert_fz_bnn}
	\end{subfigure}
	\caption{BNN: Uncertainty Plots of: a) $F_x$, b) $F_y$, c) $F_z$ for Fixed Wing Aircraft Flight Data}
	\label{fig:uncert_monty_bnn}
\end{figure}

%%%%%%%%%%%%%%%%%%%%%%%%%%%%%%%%%%%%%%%%%%%%%%%%%%%%%%%%%%%%%%%%%%%%%%
\section{Concluding Remarks}
\label{sec:conclusion}
%%%%%%%%%%%%%%%%%%%%%%%%%%%%%%%%%%%%%%%%%%%%%%%%%%%%%%%%%%%%%%%%%%%%%%

This paper explores methods for efficient quantification of modeling uncertainties in hybrid physics-informed machine learning (PIML) architectures, enabled by integrating Bayesian Neural Networks (BNNs) into them. The uncertainty propagation scheme leverages auto-differentiable physics models (within the PIML) to carry uncertainties through to the outputs. The BNN models used here was limited to probabilistic weights to the output layer, so as to keep training performance tractable. To address the sensitivity of BNNs to prior distributions, a two-stage training scheme was introduced, where, first, a PIML-ANN was trained, and its learned weights were then used as priors for the BNN, followed by the PIML-BNN training. The framework is evaluated on two case studies -- an analytical benchmark function and aerodynamics modeling for a fixed-wing RC aircraft with flight experiment data. In both cases, the modeling accuracy of the PIML-BNN was slightly poorer comparable to the other baselines such as the ANN, BNN and PIML-ANN, while still being better than the purely partial physics model. However, for uncertainty quantification, end-to-end Monte Carlo simulations, where weights are sampled and propagated through the entire physics model, demonstrated better performance than approaches that rely on Monte Carlo sampling within the BNN followed by uncertainty propagation via Taylor series approximation.

Future work could focus on integrating alternative probabilistic ML models such as Gaussian Processes and epistemic ANNs into PIML architectures and compare/contrast with PIML-BNN, especially to enable single-stage training and thus reduce overall training costs. In addition, future extension of the PIML architectures to consider input uncertainty along with model-induced uncertainty (currently considered) will significantly expand the usability of the UQ-capable PIML architectures in various robust decision-making processes.

\section*{Acknowledgements}
This work was supported by a sub-contract from Bechamo LLC under the NASA Phase II \textit{Small Business Innovation Research (SBIR)} Award No. 80NSSC22CA046. Any opinions, findings, conclusions, or recommendations expressed in this paper are those of the authors and do not necessarily reflect the views of NASA.

%%%%%%%%%%%%%%%%%%%%%%%%%%%%%%%%%%%%%%%%%%%%%%%%%%%%%%%%%%%%%%%%%%%%%%
% The bibliography is stored in an external database file
% in the BibTeX format (file_name.bib).  The bibliography is
% created by the following command and it will appear in this
% position in the document. You may, of course, create your
% own bibliography by using thebibliography environment as in
%
% \begin{thebibliography}{12}
% ...
% \bibitem{itemreference} D. E. Knudsen.
% {\em 1966 World Bnus Almanac.}
% {Permafrost Press, Novosibirsk.}
% ...
% \end{thebibliography}

% Here's where you specify the bibliography database file.
% The full file name of the bibliography database for this
% article is asme2e.bib. The name for your database is up
% to you.
% \bibliographystyle{asmems4}
% \bibliographystyle{unsrt}
% \bibliography{asme2e}
% \printbibliography

\printbibliography

@article{kendall,
  title={What uncertainties do we need in bayesian deep learning for computer vision?},
  author={Kendall, Alex and Gal, Yarin},
  journal={Advances in neural information processing systems},
  volume={30},
  year={2017}
}

@article{shahriari,
  title={Taking the human out of the loop: A review of Bayesian optimization},
  author={Shahriari, Bobak and Swersky, Kevin and Wang, Ziyu and Adams, Ryan P and De Freitas, Nando},
  journal={Proceedings of the IEEE},
  volume={104},
  number={1},
  pages={148--175},
  year={2015},
  publisher={IEEE}
}

@article{depeweg,
  title={Uncertainty decomposition in bayesian neural networks with latent variables},
  author={Depeweg, Stefan and Hern{\'a}ndez-Lobato, Jos{\'e} Miguel and Doshi-Velez, Finale and Udluft, Steffen},
  journal={arXiv preprint arXiv:1706.08495},
  year={2017}
}

@article{green,
  title={Bayesian and Markov chain Monte Carlo methods for identifying nonlinear systems in the presence of uncertainty},
  author={Green, Peter L and Worden, Keith},
  journal={Philosophical Transactions of the Royal Society A: Mathematical, Physical and Engineering Sciences},
  volume={373},
  number={2051},
  pages={20140405},
  year={2015},
  publisher={The Royal Society Publishing}
}

@article{goan,
  title={Bayesian neural networks: An introduction and survey},
  author={Goan, Ethan and Fookes, Clinton},
  journal={Case Studies in Applied Bayesian Data Science: CIRM Jean-Morlet Chair, Fall 2018},
  pages={45--87},
  year={2020},
  publisher={Springer}
}

@article{chen2002interval,
  title={Interval static displacement analysis for structures with interval parameters},
  author={Chen, Suhuan and Lian, Huadong and Yang, Xiaowei},
  journal={International Journal for Numerical Methods in Engineering},
  volume={53},
  number={2},
  pages={393--407},
  year={2002},
  publisher={Wiley Online Library}
}

@article{degrauwe,
  title={Improving interval analysis in finite element calculations by means of affine arithmetic},
  author={Degrauwe, Daan and Lombaert, Geert and De Roeck, Guido},
  journal={Computers \& structures},
  volume={88},
  number={3-4},
  pages={247--254},
  year={2010},
  publisher={Elsevier}
}

@article{abdo2016uncertainty,
  title={Uncertainty quantification in dynamic system risk assessment: a new approach with randomness and fuzzy theory},
  author={Abdo, Houssein and Flaus, Jean-Marie},
  journal={International Journal of Production Research},
  volume={54},
  number={19},
  pages={5862--5885},
  year={2016},
  publisher={Taylor \& Francis}
}

@article{zhang2021modern,
  title={Modern Monte Carlo methods for efficient uncertainty quantification and propagation: A survey},
  author={Zhang, Jiaxin},
  journal={Wiley Interdisciplinary Reviews: Computational Statistics},
  volume={13},
  number={5},
  pages={e1539},
  year={2021},
  publisher={Wiley Online Library}
}

@article{olivier2021bayesian,
  title={Bayesian neural networks for uncertainty quantification in data-driven materials modeling},
  author={Olivier, Audrey and Shields, Michael D and Graham-Brady, Lori},
  journal={Computer methods in applied mechanics and engineering},
  volume={386},
  pages={114079},
  year={2021},
  publisher={Elsevier}
}

@article{farid2022data,
  title={Data-driven method for real-time prediction and uncertainty quantification of fatigue failure under stochastic loading using artificial neural networks and Gaussian process regression},
  author={Farid, Maor},
  journal={International Journal of Fatigue},
  volume={155},
  pages={106415},
  year={2022},
  publisher={Elsevier}
}

@article{ceccarelli2021bayesian,
  title={Bayesian physics-informed neural networks for inverse uncertainty quantification problems in cardiac electrophysiology},
  author={Ceccarelli, Daniele},
  year={2021},
  publisher={Italy}
}

@article{molnar2022flow,
  title={Flow field tomography with uncertainty quantification using a Bayesian physics-informed neural network},
  author={Molnar, Joseph P and Grauer, Samuel J},
  journal={Measurement Science and Technology},
  volume={33},
  number={6},
  pages={065305},
  year={2022},
  publisher={IOP Publishing}
}

@article{pfortner2022physics,
  title={Physics-informed Gaussian process regression generalizes linear PDE solvers},
  author={Pf{\"o}rtner, Marvin and Steinwart, Ingo and Hennig, Philipp and Wenger, Jonathan},
  journal={arXiv preprint arXiv:2212.12474},
  year={2022}
}

@article{mumpower2022physically,
  title={Physically interpretable machine learning for nuclear masses},
  author={Mumpower, MR and Sprouse, TM and Lovell, AE and Mohan, AT},
  journal={Physical Review C},
  volume={106},
  number={2},
  pages={L021301},
  year={2022},
  publisher={APS}
}

@article{mahadevan2022uncertainty,
  title={Uncertainty quantification for additive manufacturing process improvement: Recent advances},
  author={Mahadevan, Sankaran and Nath, Paromita and Hu, Zhen},
  journal={ASCE-ASME Journal of Risk and Uncertainty in Engineering Systems, Part B: Mechanical Engineering},
  volume={8},
  number={1},
  pages={010801},
  year={2022},
  publisher={American Society of Mechanical Engineers}
}

@article{kapusuzoglu2021information,
  title={Information fusion and machine learning for sensitivity analysis using physics knowledge and experimental data},
  author={Kapusuzoglu, Berkcan and Mahadevan, Sankaran},
  journal={Reliability Engineering \& System Safety},
  volume={214},
  pages={107712},
  year={2021},
  publisher={Elsevier}
}

@article{hao2023physics,
  title={A physics-informed machine learning approach for notch fatigue evaluation of alloys used in aerospace},
  author={Hao, WQ and Tan, L and Yang, XG and Shi, DQ and Wang, ML and Miao, GL and Fan, YS},
  journal={International Journal of Fatigue},
  volume={170},
  pages={107536},
  year={2023},
  publisher={Elsevier}
}

@article{xu2023physics,
  title={Physics-informed machine learning for reliability and systems safety applications: State of the art and challenges},
  author={Xu, Yanwen and Kohtz, Sara and Boakye, Jessica and Gardoni, Paolo and Wang, Pingfeng},
  journal={Reliability Engineering \& System Safety},
  volume={230},
  pages={108900},
  year={2023},
  publisher={Elsevier}
}

@book{fuks2020physics,
  title={Physics Informed Machine Learning and Uncertainty Propagation for Multiphase Transport in Porous Media},
  author={Fuks, Olga},
  year={2020},
  publisher={Stanford University}
}

@article{GAO2023107361,
title = {Multiaxial fatigue prediction and uncertainty quantification based on back propagation neural network and Gaussian process regression},
journal = {International Journal of Fatigue},
volume = {168},
pages = {107361},
year = {2023},
issn = {0142-1123},
doi = {https://doi.org/10.1016/j.ijfatigue.2022.107361},
url = {https://www.sciencedirect.com/science/article/pii/S0142112322006119},
author = {Jingjing Gao and Jun Wang and Zili Xu and Cunjun Wang and Song Yan}
}

@InProceedings{	  cheng2009fusion,
  title		= {A fusion prognostics method for remaining useful life
		  prediction of electronic products},
  author	= {Cheng, Shunfeng and Pecht, Michael},
  booktitle	= {Automation Science and Engineering, 2009. CASE 2009. IEEE
		  International Conference on},
  pages		= {102--107},
  year		= {2009},
  organization	= {IEEE}
}

@Article{	  jagtap2020adaptive,
  title		= {Adaptive activation functions accelerate convergence in
		  deep and physics-informed neural networks},
  author	= {Jagtap, Ameya D and Kawaguchi, Kenji and Karniadakis,
		  George Em},
  journal	= {Journal of Computational Physics},
  volume	= {404},
  pages		= {109136},
  year		= {2020},
  publisher	= {Elsevier}
}

@PhDThesis{	  javed2014robust,
  title		= {A robust \& reliable Data-driven prognostics approach
		  based on extreme learning machine and fuzzy clustering.},
  author	= {Javed, Kamran},
  year		= {2014},
  school	= {Universit{\'e} de Franche-Comt{\'e}}
}

@Article{	  karniadakis2021physics,
  title		= {Physics-informed machine learning},
  author	= {Karniadakis, George Em and Kevrekidis, Ioannis G and Lu,
		  Lu and Perdikaris, Paris and Wang, Sifan and Yang, Liu},
  journal	= {Nature Reviews Physics},
  volume	= {3},
  number	= {6},
  pages		= {422--440},
  year		= {2021},
  publisher	= {Nature Publishing Group}
}

@Article{	  karpatne2017physics,
  title		= {Physics-guided Neural Networks (PGNN): An Application in
		  Lake Temperature Modeling},
  author	= {Karpatne, Anuj and Watkins, William and Read, Jordan and
		  Kumar, Vipin},
  journal	= {arXiv preprint arXiv:1710.11431},
  year		= {2017}
}

@Article{	  li2020fourier,
  title		= {Fourier neural operator for parametric partial
		  differential equations},
  author	= {Li, Zongyi and Kovachki, Nikola and Azizzadenesheli,
		  Kamyar and Liu, Burigede and Bhattacharya, Kaushik and
		  Stuart, Andrew and Anandkumar, Anima},
  journal	= {arXiv preprint arXiv:2010.08895},
  year		= {2020}
}

@Article{	  lu2021learning,
  title		= {Learning nonlinear operators via DeepONet based on the
		  universal approximation theorem of operators},
  author	= {Lu, Lu and Jin, Pengzhan and Pang, Guofei and Zhang,
		  Zhongqiang and Karniadakis, George Em},
  journal	= {Nature Machine Intelligence},
  volume	= {3},
  number	= {3},
  pages		= {218--229},
  year		= {2021},
  publisher	= {Nature Publishing Group}
}

@Article{	  nourani2009combined,
  title		= {A combined neural-wavelet model for prediction of
		  Ligvanchai watershed precipitation},
  author	= {Nourani, Vahid and Alami, Mohammad T and Aminfar, Mohammad
		  H},
  journal	= {Engineering Applications of Artificial Intelligence},
  volume	= {22},
  number	= {3},
  pages		= {466--472},
  year		= {2009},
  publisher	= {Elsevier}
}

@Article{	  rai2020driven,
  title		= {Driven by data or derived through physics? a review of
		  hybrid physics guided machine learning techniques with
		  cyber-physical system (cps) focus},
  author	= {Rai, Rahul and Sahu, Chandan K},
  journal	= {IEEE Access},
  volume	= {8},
  pages		= {71050--71073},
  year		= {2020},
  publisher	= {IEEE}
}

@Article{	  raissi2019physics,
  title		= {Physics-informed neural networks: A deep learning
		  framework for solving forward and inverse problems
		  involving nonlinear partial differential equations},
  author	= {Raissi, Maziar and Perdikaris, Paris and Karniadakis,
		  George E},
  journal	= {Journal of Computational Physics},
  volume	= {378},
  pages		= {686--707},
  year		= {2019},
  publisher	= {Elsevier}
}

@InProceedings{	  singh2019pi,
  title		= {PI-LSTM: Physics-Infused Long Short-Term Memory Network},
  author	= {Singh, Shubhendu Kumar and Yang, Ruoyu and Behjat, Amir
		  and Rai, Rahul and Chowdhury, Souma and Matei, Ion},
  booktitle	= {2019 18th IEEE International Conference On Machine
		  Learning And Applications (ICMLA)},
  pages		= {34--41},
  year		= {2019},
  organization	= {IEEE}
}

@Article{	  young2017physically,
  title		= {A physically based and machine learning hybrid approach
		  for accurate rainfall-runoff modeling during extreme
		  typhoon events},
  author	= {Young, Chih-Chieh and Liu, Wen-Cheng and Wu, Ming-Chang},
  journal	= {Applied Soft Computing},
  volume	= {53},
  pages		= {205--216},
  year		= {2017},
  publisher	= {Elsevier}
}

@InProceedings{	  yang2017investigating,
  title		= {Investigating grey-box modeling for predictive analytics
		  in smart manufacturing},
  author	= {Yang, Zhuo and Eddy, Douglas and Krishnamurty, Sundar and
		  Grosse, Ian and Denno, Peter and Lu, Yan and Witherell,
		  Paul},
  booktitle	= {International design engineering technical conferences and
		  computers and information in engineering conference},
  volume	= {58134},
  pages		= {V02BT03A024},
  year		= {2017},
  organization	= {American Society of Mechanical Engineers}
}

@Article{	  manoharan2019grey,
  title		= {A grey box software framework for sustainability
		  assessment of composed manufacturing processes: A hybrid
		  manufacturing case},
  author	= {Manoharan, Sriram and Haapala, Karl R},
  journal	= {Procedia CIRP},
  volume	= {80},
  pages		= {440--445},
  year		= {2019},
  publisher	= {Elsevier}
}

@Article{	  mao2020physics,
  title		= {Physics-informed neural networks for high-speed flows},
  author	= {Mao, Zhiping and Jagtap, Ameya D and Karniadakis, George
		  Em},
  journal	= {Computer Methods in Applied Mechanics and Engineering},
  volume	= {360},
  pages		= {112789},
  year		= {2020},
  publisher	= {Elsevier}
}

@Article{	  zhang2022analyses,
  title		= {Analyses of internal structures and defects in materials
		  using physics-informed neural networks},
  author	= {Zhang, Enrui and Dao, Ming and Karniadakis, George Em and
		  Suresh, Subra},
  journal	= {Science advances},
  volume	= {8},
  number	= {7},
  pages		= {eabk0644},
  year		= {2022},
  publisher	= {American Association for the Advancement of Science}
}

@Article{	  cai2021physics,
  title		= {Physics-informed neural networks (PINNs) for fluid
		  mechanics: A review},
  author	= {Cai, Shengze and Mao, Zhiping and Wang, Zhicheng and Yin,
		  Minglang and Karniadakis, George Em},
  journal	= {Acta Mechanica Sinica},
  volume	= {37},
  number	= {12},
  pages		= {1727--1738},
  year		= {2021},
  publisher	= {Springer}
}

@Article{	  faroughi2023physics,
  title		= {Physics-informed neural networks with periodic activation
		  functions for solute transport in heterogeneous porous
		  media},
  author	= {Faroughi, Salah A and Soltanmohammadi, Ramin and Datta,
		  Pingki and Mahjour, Seyed Kourosh and Faroughi, Shirko},
  journal	= {Mathematics},
  volume	= {12},
  number	= {1},
  pages		= {63},
  year		= {2023},
  publisher	= {MDPI}
}

@Article{	  cuomo2022scientific,
  title		= {Scientific machine learning through physics--informed
		  neural networks: Where we are and what’s next},
  author	= {Cuomo, Salvatore and Di Cola, Vincenzo Schiano and
		  Giampaolo, Fabio and Rozza, Gianluigi and Raissi, Maziar
		  and Piccialli, Francesco},
  journal	= {Journal of Scientific Computing},
  volume	= {92},
  number	= {3},
  pages		= {88},
  year		= {2022},
  publisher	= {Springer}
}

@InProceedings{	  oddiraju2023physics,
  title		= {Physics Infused Machine Learning Based Prediction of VTOL
		  Aerodynamics with Sparse Datasets},
  author	= {Oddiraju, Manaswin and Amin, Divyang and Piedmonte,
		  Michael and Chowdhury, Souma},
  booktitle	= {AIAA AVIATION 2023 Forum},
  pages		= {4376},
  year		= {2023}
}

@article{hoffman2013stochastic,
  title={Stochastic variational inference},
  author={Hoffman, Matthew D and Blei, David M and Wang, Chong and Paisley, John},
  journal={the Journal of machine Learning research},
  volume={14},
  number={1},
  pages={1303--1347},
  year={2013},
  publisher={JMLR. org}
}

@article{oddiraju2024laser,
    author = {Oddiraju, Manaswin and Cleeman, Jeremy and Malhotra, Rajiv and Chowdhury, Souma},
    title = {A Differentiable Physics-Informed Machine Learning Approach to Model Laser-based Micro-Manufacturing Process},
    journal = {Journal of Manufacturing Science and Engineering},
    pages = {1-13},
    year = {2024},
    month = {12},
    issn = {1087-1357},
    doi = {10.1115/1.4067355},
    url = {https://doi.org/10.1115/1.4067355},
}

@Inproceedings{oddiraju2024physics,
  title={Physics-Informed Machine Learning Towards A Real-Time Spacecraft Thermal Simulator},
  author={Oddiraju, Manaswin and Hasnain, Zaki and Bandyopadhyay, Saptarshi and Sunada, Eric and Chowdhury, Souma},
  booktitle={AIAA AVIATION FORUM AND ASCEND 2024},
  pages={4204},
  year={2024}
}
%%%%%%%%%%%%%%%%%%%%%%%%%%%%%%%%%%%%%%%%%%%%%%%%%%%%%%%%%%%%%%%%%%%%%%
\appendix       %%% starting appendix
\section*{Appendix A: KL-divergence and ELBO}\label{appendix}
To get the intractable posterior distribution in variational inference, we first define a parametrized and tractable stand-in distribution, called the approximate posterior $q_\phi(w)$, then tune the parameters $\phi$ so that it better approximates the intractable distribution. Developing a variational method for approximate inference requires two steps:\\
1. Formalizing a notion of similarity between two probability distributions\\
2. Writing down a tractable optimization problem that corresponds to maximizing this notion of similarity\\
We use the Kullback-Liebler divergence (or KL-divergence) to convey dissimilarity between two distributions. Expression for KL-divergence between approximate posterior ($q_\phi(w)$) and true posterior ($p(w \mid D)$) is:
\begin{equation}
	\begin{aligned}
		d_{K L}\left[q_\phi(w) \| p(w \mid D)\right] & =\int_w q_\phi(w) \log \frac{q_\phi(w)}{p(w \mid D)}                  \\
		                                             & =\mathbb{E}_{q_\theta(w)}\left[\log q_\phi(w)-\log p(w \mid D)\right] \\
		                                             & =\mathbb{E}_{q_\phi(w)}\left[\log q_\phi(w)-\log p(w, D)\right]       \\
		                                             & \;\;\;\;\;\;\;\;\;\;\;\;\;\;\;\;          +\log p(D)
	\end{aligned}
\end{equation}
Taking the expectation of the remaining terms over the data distribution $D$ yields a tractable and suitable surrogate objective called the Evidence Lower Bound (ELBO):

\begin{equation}
	\begin{aligned}
		\tilde{L}(\phi) & =\mathbb{E}_{D, q_\phi(w)}\left[\log p(w, D)-\log q_\phi(w)\right]                  \\
		                & =\mathbb{E}_{x, y \sim D}[\mathbb{E}_{w \sim q_\phi(w)}[\log p(\hat{y}(x)=y \mid w) \\
		                & \;\;\;\;\;\;\;\;\;\;\;\;\;\;+\log p(w)-\log q_\phi(w)]]
	\end{aligned}
\end{equation}
Our goal is to maximize ELBO to get the posterior distribution of weights.

$$
	\underset{\phi}{\operatorname{maximize}} \;\;\tilde{L}(\phi)
$$
%%%%%%%%%%%%%%%%%%%%%%%%%%%%%%%%%%%%%%%%%%%%%%%%%%%%%%%%%%%%%%%%%%%%%%
\section*{Appendix B: Convergence Plots}
\subsection*{Gramacy \& Lee Problem}
Fig. \ref{fig:train_gl} and \ref{fig:test_gl} show the convergence history of the Gramacy $\&$ Lee problem. From the convergence plots, it is clear that models with the physics model are struggling to converge. This is due to the high number of hidden dimensions in the transfer layer for a simple problem owing to requiring more training.

\begin{figure*}
	\centering
	\begin{subfigure}{0.45\linewidth}
		\centering
		\includegraphics[width=\linewidth]{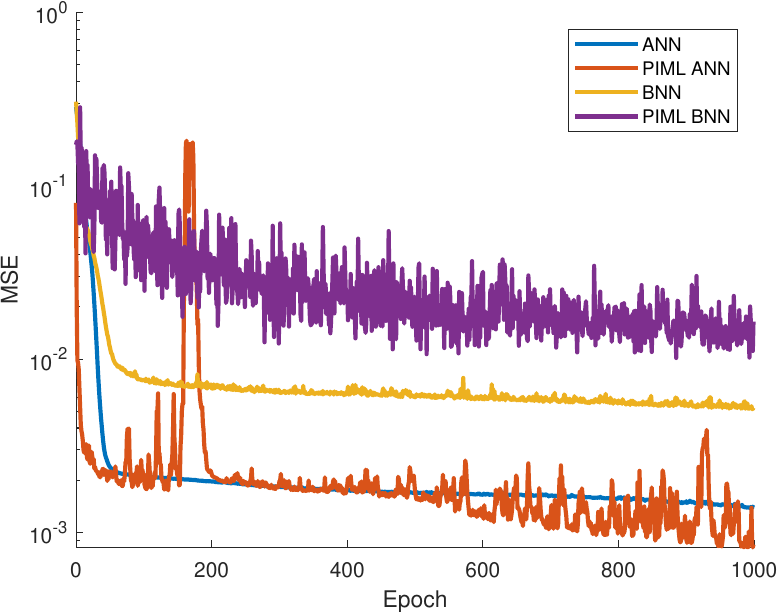}
		\caption{}
		\label{fig:train_gl}
	\end{subfigure}
	\begin{subfigure}{0.45\linewidth}
		\centering
		\includegraphics[width=\linewidth]{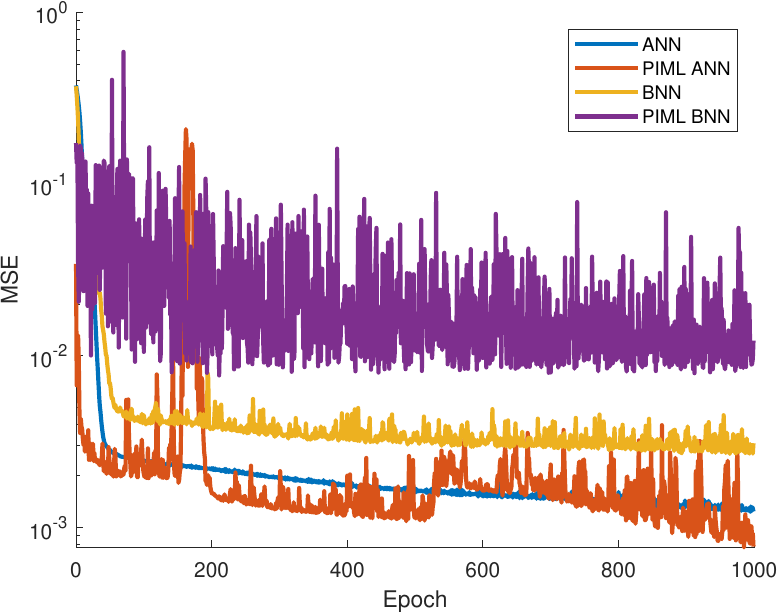}
		\caption{}
		\label{fig:test_gl}
	\end{subfigure}
	\caption{Convergence History of Models Training on Gramacy $\&$ Lee Problem: a) Train loss, b) Test Loss}
	\label{fig:conv_gl}
\end{figure*}
From Fig. \ref{fig:pred_nn},  we can see clearly that PIML-ANN is predicting better compared to pure data-driven ANN even though it is struggling to converge as smoothly as pure data-driven ANN.

\subsection*{Aerodynamics of Fixed-Wing Aircraft}
Fig. \ref{fig:train_monty} and \ref{fig:test_monty} show the convergence history of the Monty Flight. From the convergence plots, we can see that all the models are training well and are converging.\\

\begin{figure*}
	\centering
	\begin{subfigure}{0.45\linewidth}
		\centering
		\includegraphics[width=\linewidth]{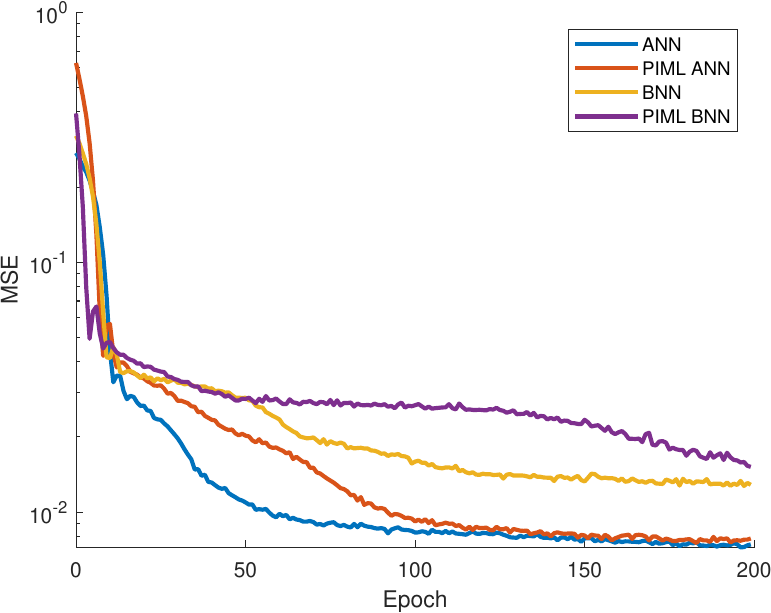}
		\caption{}
		\label{fig:train_monty}
	\end{subfigure}
	\begin{subfigure}{0.45\linewidth}
		\centering
		\includegraphics[width=\linewidth]{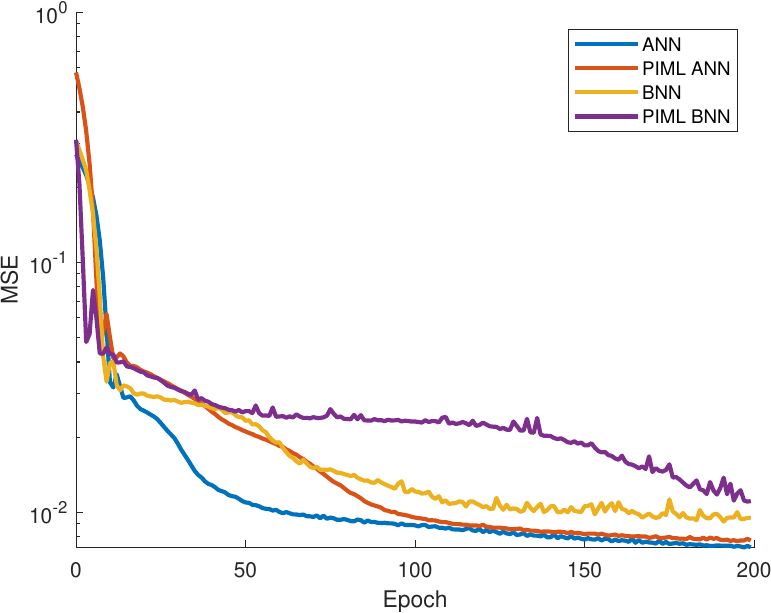}
		\caption{}
		\label{fig:test_monty}
	\end{subfigure}
	\caption{Convergence History of Models Training on Fixed Wing Aircraft Flight Data: a) Train loss, b) Test Loss}
	\label{fig:conv_monty}
\end{figure*}

\end{document}